\documentclass{article}

\usepackage[numbers, sort]{natbib}

\usepackage{graphicx}
\usepackage{wrapfig}
\usepackage{amsmath}

\usepackage[preprint]{HEAP}

\usepackage[utf8]{inputenc} 
\usepackage[T1]{fontenc}    
\usepackage{hyperref}       
\usepackage{url}            
\usepackage{booktabs}       
\usepackage{amsfonts}       
\usepackage{nicefrac}       
\usepackage{microtype}      
\usepackage{xcolor}         

\usepackage{comment}

\title{Hierarchical-embedding autoencoder with a predictor (HEAP) as efficient architecture for learning long-term evolution of complex multi-scale physical systems}

\author{%
  Alexander I.~Khrabry \\
  Princeton University\\
  Princeton, NJ 08544 \\
  \texttt{akhrabry@princeton.edu} \\
  \And
  Edward A. Startsev \\
  Princeton Plasma Physics Laboratory \\
  Princeton, NJ 08540 \\
  \texttt{estarts@pppl.gov} \\
  \AND
  Andrew T. Powis \\
  Princeton University\\
  Princeton, NJ 08544 \\
  \texttt{apowis@princeton.edu} \\
   \And
   Igor D. Kaganovich \\
   Princeton Plasma Physics Laboratory \\
   Princeton, NJ 08540 \\
   \texttt{ikaganov@pppl.gov} \\
}

\begin{document}

\maketitle

\begin{abstract}
  We propose a novel efficient architecture for learning long-term evolution in complex multi-scale physical systems which is based on the idea of separation of scales. Structures of various scales that dynamically emerge in the system interact with each other only locally. Structures of similar scale can interact directly when they are in contact and indirectly when they are parts of larger structures that interact directly. This enables modeling a multi-scale system in an efficient way, where interactions between small-scale features that are apart from each other do not need to be modeled. The hierarchical fully-convolutional autoencoder transforms the state of a physical system not just into a single embedding layer, as it is done conventionally, but into a series of embedding layers which encode structures of various scales preserving spatial information at a corresponding resolution level. Shallower layers embed smaller structures on a finer grid, while deeper layers embed larger structures on a coarser grid. The predictor advances all embedding layers in sync. Interactions between features of various scales are modeled using a combination of convolutional operators. We compare the performance of our model to variations of a conventional ResNet architecture in application to the Hasegawa-Wakatani turbulence. A multifold improvement in long-term prediction accuracy was observed for crucial statistical characteristics of this system.
\end{abstract}

\section{Introduction}\label{Introduction}

Predicting long-term evolution in complex multi-scale physical systems is crucial in many fields (high Reynolds number liquid and gas flows \cite{W98}, fully and partially magnetized plasmas \cite{H12}, weather forecasting \cite{Lam3} etc.). The dynamics of such systems are often characterized by emerging structures (vortices and waves) of various scale. In bounded systems, the largest structures typically scale with the size of the system, while the scale of the smallest structures (the Kolmogorov scale \cite{K61}), is often orders of magnitude smaller. Long-range interactions between emerging structures can take place due to high speed of information propagation in the system. For example, in plasma/gas/liquid flows, acoustic waves propagate at the speed of sound or, in the case of incompressible fluids, at infinite speed. This multi-scale nature of emergent features and interactions between them makes numerical simulation difficult and expensive, since spatial and temporal resolution must be high enough to resolve features of all scales. At the same time, implicit numerical schemes are required to solve systems with high information propagation speed. These challenges create a strong demand for machine learning-accelerated predictions of multi-scale systems. 

In this context, machine learning has been applied in various capacities \cite{W4,L23,V2}: (1) solver acceleration with a neural-network-based sub-scale model \cite{U1, K1, G3, A0, B19, L19, S1, OS20} or learned initial conditions \cite{Tasman}, (2) training a surrogate model to replace the solver \cite{S19, B3, A4, Z0, Lu1, W1}. The latter approach can be further divided into two broad categories: physics-informed methods that incorporate knowledge of the underlying PDE in the loss function \cite{R19, K21, D4, Z1, Z3, J1, E2}, or purely data-driven methods \cite{C4, S2, W20, G4, L3, Z4, K19}. We focus on the latter since the underlying PDE is not always known, e.g., when training from experimental observations, or when a reduced order surrogate model is required (e.g., a fluid surrogate model based on kinetic modeling data \cite{La22}). Also, oftentimes, a surrogate model is desired that uses low temporal or spatial resolution data \cite{S2} for which the underlying PDE (even if known) may not be numerically satisfied.

{\bf Our main contributions are:}

•	We present the Hierarchical-Embedding Autoencoder with a Predictor (HEAP) architecture. Predictions are made in the embedding space, which, unlike conventional models, has a hierarchical (layered) structure. This architecture enables modeling a multi-scale system in an efficient way: with a small model, low computational cost, and small amount of training data.

•	We apply the HEAP architecture to learn the temporal evolution of Hasegawa-Wakatani (HW) magnetized plasma turbulence as a case study and demonstrate its ability to accurately predict key statistical characteristics of this system over long model rollouts.

•	We show substantial improvement in accuracy over conventional architectures.

{\bf Applicability and limitations of the study:}

We test our architecture using a case study of Hasegawa-Wakatani plasma turbulence \cite{H83}, which is a model for highly magnetized plasmas typical in fusion plasma devices (see Appendix \ref{A_HW}). This model is commonly used as a test bed for ML approaches \cite{C4, G4, U1, H0}, and we expect the results to be applicable to other continuous multi-scale physical systems.

We study a 2D system, but the model can be extended to 3D systems using 3D convolutions. We work with structured grids, which allow us to apply convolutions. However, this approach can be adapted to unstructured data using graph networks (GNNs) \cite{P1, K5, G22, G23, F2}, as demonstrated in U-Net GNNs \cite{J4, D24}, by utilizing graph convolutions.

\section{Related work}

Several architectures have been used to construct surrogate models for complex multi-scale physical systems. One line of work is based on the transformer architecture \cite{V17} adapted for 2D/3D data, i.e., a vision transformer (ViT \cite{D0}) \cite{M4, C022, C0, Z2, N3, A4}. ViT creates individual embeddings for patches that partition the system (typically as equally sized squares). Interactions of all length scales between the patches are accounted for through the self-attention mechanism where each patch attends to all other patches. However, this capability comes at a cost of quadratic computational complexity in respect to the system size (i.e., area for 2D systems) and higher requirements for the amount of training data. A number of work aim at reducing transformer complexity to linear for the purpose of physics modeling \cite{L023, Cao1, Hao3, Li3}, however, this typically comes at the price of reducing accuracy to levels which can already be achieved with alternative models. An additional challenge with transformer models is the need for learnable relative (periodic) positional embeddings \cite{S18, Wu21} to handle periodic data, such as in Hasegawa-Wakatani (WH) turbulence, which further complicates the model. 

Another line of work considers learning the dynamics of a Fourier space representation. Fourier Neural Operators \cite{P2, T3, He3, Z0, Z0, R2} (FNOs), naturally handle multiscale interactions through parameterization of various Fourier modes. However, they can struggle with capturing fine-scale details and local spatial dependencies \cite{L5}.

A third direction of research focuses on using various convolutional architectures for regular grid data and graph neural network (GNNs) architectures for irregular grids. Convolutions on regular grids and message passing in GNNs \cite{B2, P1, SG20} are efficient at capturing local interactions (between neighbor embeddings). Several approaches in the literature aim to incorporate longer-range interactions into the model, many of which utilize the Encode-Process-Decode paradigm \cite{B18, SG18, SG20}. As the name suggests, an encoder first creates embeddings $y_i$ of the input data $x_i$ (i.e., 2D or 3D fields) at a time step $i$. Typically, the data is down-sampled in physical space, while the depth of the feature map (i.e., the number of channels) is increased. The processor (or predictor) then advances the embeddings $y_i$ one time step forward (to $\hat{y}_{i+1}$). Finally, the decoder unpacks and up-samples the embedded data $\hat{y}_{i+1}$ back to physical space, yielding $\hat{x}_{i+1}$. This algorithm is applied auto-regressively to generate solutions for a sequence of time steps \{$\hat{x}_{i+1}$, $\hat{x}_{i+2}$, …\}. Multiple variations exist in how this paradigm is applied.

When the input data is down-sampled, embedding units are combined into larger units with the following implications. 1) The interaction distance in physical space increases since some embedding units that were previously separated are now merged into units that are neighbors, which can interact through convolutions/message passing. 2) An explicit representation of spatial information is lost for the merged units. The problem of losing spatial information is partially alleviated by U-Nets \cite{R15, D24, J4, W20} and DenseNets \cite{GH16, J17}, where lateral skip connections are placed between embeddings of a down-sampling path and embeddings of an up-sampling path of the same level (same spatial resolution). U-net/DenseNet can either resemble all the parts of the encoder-predictor-decoder framework (with the down-sampling path as the encoder and the up-sampling path as the decoder) \cite{M2, EP4}, or be used solely as a predictor alongside a convolutional encoder and decoder \cite{S2}. One weakness of a U-Net as a predictor is the misalignment between embeddings of up-sampling and down-sampling paths \cite{Z4} which occurs because these embeddings are shifted in time. To address this, in SineNet \cite{Z4} several U-Nets are stacked sequentially \cite{X17, So18} to reduce a time step (and thereby reduce misalignment) in each.

Another popular choice is the use of a ResNet \cite{H16} as a predictor in combination with a convolutional encoder and decoder \cite{C21, S2, K19}. ResNet predictors typically operate at a single spatial resolution level by using flat (stride=1) convolutions and thereby do not have a misalignment issue. However, only local interactions between neighbor embeddings are modeled. A modification of Resnet, DilResNet \cite{S2} uses dilated convolutions to account for long-range interactions. Notably, dilated convolutions (especially with high dilation ratios of 4 and 8 that are used in DilResNet) take longer to compute compared to regular convolutions. The model size in DilResNet needs to be an order of magnitude smaller compared to a ResNet or a SineNet to achieve a similar training time \cite{Z4}.

Yet another variation of the encoder-predictor-decoder paradigm is implemented in MeshGraphNets \cite{F2, G23}, where two predictors are applied in parallel at two resolution levels in a graph, thereby synchronizing all layers in time and avoiding the misalignment problem.

Multiple published tests \cite{S2, Z4, L023, G2} indicate that U-Net, ResNet, DilResNet and linear attention transformers operate at a similar level of accuracy with a slight advantage of one or another model (typically measured by percent or tens of percent difference) depending on the physical system modeled. FNO models showed similar to slightly worse accuracy.

Another crucial aspect in which the encoder-predictor-decoder architectures vary is whether all three components are trained together to make predictions in the physical space (predict $\hat{x}_{i+1}$ from $x_i$, as in U-Net based architectures and \cite{S2, G23}) or the encoder-decoder are trained separately as an autoencoder \cite{H23, C22, E0, C4, K19}. In the latter case, multi-time-step predictions are made auto-regressively in the embedding space ($\hat{y}_{i+1}$ from $y_i$, $\hat{y}_{i+2}$ from $\hat{y}_{i+1}$ etc.)\footnote{RNN \cite{C4} or LSTM \cite{E0} networks can be used to capture long-term temporal correlations in the system. This is efficient when a model misses some part of the physics, e.g., in \cite{C4} only a field of $\phi$ was modeled where the full physics is described by two 2D fields, $\phi$ and $n$. When all physics is included in the model, one-step prediction should be sufficient and significantly computationally cheaper.} directly using embedding from the previous step, avoiding the need to decode and re-encode them at each time step as in the former case. This is commonly referred to as a reduced-order-model (ROM) approach; it seems to be advantageous for prediction accuracy \cite{E0, C22, G22}. However, in this approach, it is difficult to balance the depth of the encoder such that spatial information is not overly degraded and long-range interactions are correctly accounted for. 

{\bf Our architecture naturally incorporates the scale hierarchy in its design} which makes it efficient at capturing physics with a cascade of length (energy) scales. It has a hierarchical autoencoder producing hierarchical embeddings of various scale and corresponding spatial resolution. Predictions are made in the embedding space, at all resolution levels, synchronized in time.

\section{The proposed architecture (HEAP)} \label{Architecture}

Our model is based on the following physical intuition. Structures of various scale emerge in complex systems. Larger structures encompass smaller ones, forming a cascade of scales and a corresponding energy cascade. Structures of similar size can interact (exchange energy, mass and momentum) directly if they are in physical contact. Structures of different scales can interact if the smaller structure is part of the larger one. Smaller structures which are apart from each other do not interact directly, yet can interact indirectly through contacting larger-scale structures which encompass them. 

This intuition implies that a state of a physical system can be encoded in a hierarchical (layered) way, where each layer contains local encodings of structures of corresponding scale. Larger structures are encoded in deeper layers of the model. Their embeddings lack small-scale details, which, in turn, are represented by embeddings of small-scale features residing in shallower layers. Interactions between the embeddings only occur locally, between neighbors within each layer (laterally) and between the layers (vertically). This enables shift-equivariance through weight-sharing to minimize the model size. The number of channels in local embedding is the same within each layer and is also the same, or changes weakly, between the layers, assuming that the complexity of features at each scale embedded by a corresponding layer are approximately the same. This approach allows for capture of local and non-local interactions in an efficient way. Computational complexity in this model scales linearly with the system’s size (see Appendix \ref{m_complex}).

Our Hierarchical Embedding Autoencoder-Predictor model (HEAP), consists of an auto-encoder (AE) and a predictor, described in detail below. The hierarchical fully-convolutional AE transforms the state of a physical system not just into a single embedding layer, as is conventionally done, but into a series of embedding layers which encode structures of various scales, preserving spatial information at a corresponding resolution level. Shallower layers embed smaller structures on a finer grid, while deeper layers embed larger structures on a coarser grid. The predictor advances the hierarchical embedding of the system for one time step at a time (and autoregressively for multiple steps) accounting for local interactions within each hierarchical layer and between the layers. All physical information about the system is fed into the AE (i.e., both fields $n$ and $\phi$ for the HW system). Since the equations feature only first order derivatives in time, a single state of the system is sufficient to predict all future states. Thereby, no recurrent model is required. 

\subsection{Hierarchical autoencoder}\label{S_HAE}

\begin{figure}[hbt!]
  \centering
  \scalebox{1}[1]{\includegraphics[width=1.0\textwidth]{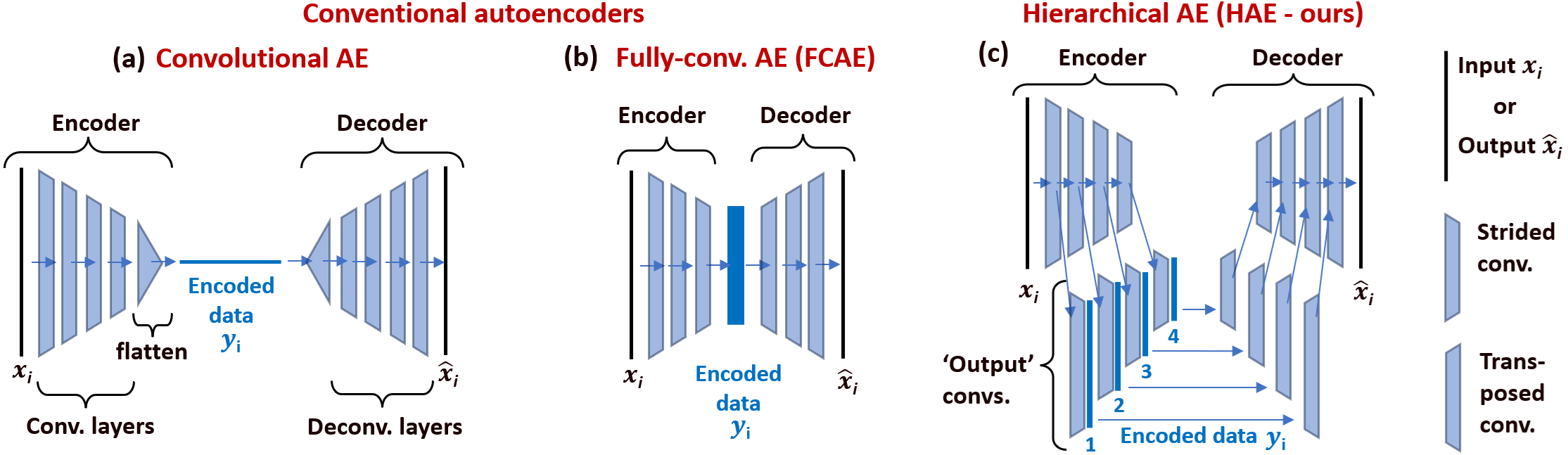}}
  \caption{Various architectures of convolutional autoencoders. Blue rectangles represent encoded data, in a single layer (a) and (b) and in multiple hierarchical layers (c). Vertical rectangles represent 2D data (on a 2D grid); a horizontal rectangle represents 1D (flattened) data. The width of vertical rectangles indicates the number of channels. Arrows represent the flow of information. }
  \label{AE}
\end{figure}

The autoencoder (AE) is used to compress the physical state data $x_i$ at every time step $i$ into a hierarchical embedding $y_i$ and recover the reconstruction of $x_i$. Our hierarchical AE (HAE, not to be confused with Stacked Convolutional Auto-Encoders \cite{M11, V8}) is illustrated in in Fig. \ref{AE}c in comparison to popular convolutional architectures with (Fig. \ref{AE}a) and without (Fig. \ref{AE}b) a fully-connected layer (fully-convolutional AE or FCAE). In our hierarchical AE, an additional strided (stride=2) ‘output’ convolution is applied at each layer of the encoder to divert and compress information from this layer into a corresponding embedding layer. Another convolution is applied to the same data to pass it ‘down’ towards deeper encoder layers. The depth number of output channels is the same for all output convolutions. It is much smaller than that in a single embedding layer produced by FCAE of similar depth, since the total number of local embeddings at all layers is considerably larger in this hierarchical model. The embeddings at each layer are passed to a corresponding layer of the decoder where they are up-sampled by a transposed convolution (deconvolution) and added to an up-sampled embedding from deeper layers. Note that our model converts into a fully-convolutional AE (FCAE) when only one layer of embeddings is used, which, in turn converts into a convolutional AE with the increase of depth, when the deepest layer has 1x1 spatial dimensions. We test how the number of hierarchical embedding layers in HAE and the model depth in a conventional FCAE affect its performance in the Experiments section \ref{Experiments}.

The number of convolutional layers presented in Fig. \ref{AE} is arbitrary (just for illustration). The details of the actual model, including layer dimensions, activation function and batch-normalization, are presented in Appendix \ref{HAE_details}, Fig. \ref{OurAE}.

\begin{figure}[hbt!]
  \centering
  \includegraphics[width=1.0\textwidth]{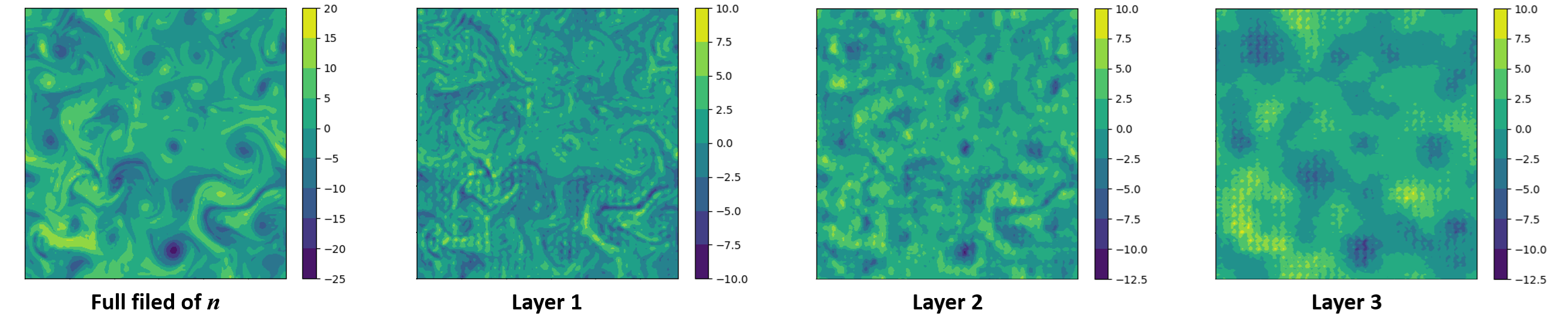}
  \caption{Multi-scale nature of structures encoded by various encoder layers. Deeper layers encode larger structures.}
  \label{mult}
\end{figure}

The multi-scale nature of structures encoded by various levels is illustrated in Fig. \ref{mult}, where the full field of density $n$ is compared to encoder outputs when only one level of embedding is used and other embedding levels are set to zero. Numbering starts from the shallowest layer which captures small scale details (see Fig. \ref{AE}). Apparently, each deeper layer encodes larger structures. Note that the sum of these outputs does not constitute the full fields due to non-linear operations within the network, these are simply given as an illustration. 

\subsection{Hierarchical predictor} \label{S_P}

\begin{figure}[hbt!]
  \centering
  \includegraphics[width=1.0\textwidth]{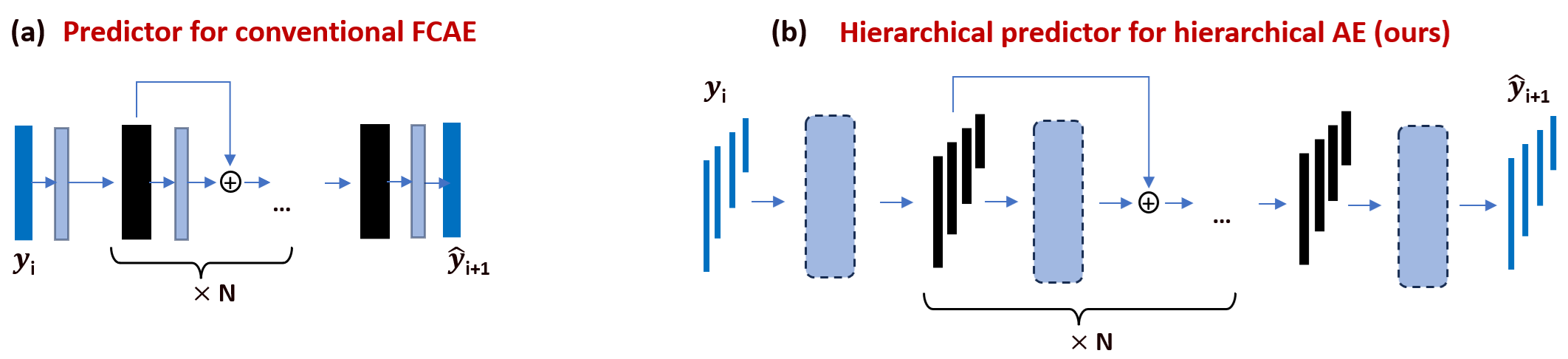}
  \caption{Predictors to couple with a conventional fully-convolutional AE (a) and hierarchical AE (b)}
  \label{P}
\end{figure}

The predictor of our model is inspired by ResNet \cite{H16}, a popular predictor choice for coupling with FCAE, see Fig. \ref{P}. The ResNet predictor consists of a series of flat convolutions. The first layer expands the number of channels (which is illustrated by thicker rectangles in Fig. \ref{P}), while the last layer compresses it back to the original value. Skip connections are added between intermediate layers (to prevent vanishing/exploding gradients). In our model, embeddings have a layered structure. We desire embeddings to interact locally, although not just within each layer, but also between various layers. To achieve this, each of the N internal steps of our predictor comprises a combination of several convolutional operators. This combination is shown schematically as a block with a dashed frame in Fig. \ref{P}b. A detailed depiction of this block and further training details are provided in Appendix \ref{P_details}. 

In other aspects, our model is similar to ResNet. We expand the depth of representation maps within each layer, use skip connections after each convolutional block, and apply layer-normalization. When the number of embedding layers in our model reduces to one, the predictor converts into a ResNet.

\section{Experiments} \label{Experiments}

We apply our model to learn the time evolution of the Hasegawa-Wakatani (HW) turbulence (see Appendix \ref{A_HW}) \cite{H83} simulated using the BOUT++ code \cite{B}. Data was collected using a fixed time step ($\Delta t = 1$) to generate solver outputs, which corresponded to hundreds to thousands of actual solver time steps. A total of output 4800 time steps were collected. The first 500 steps corresponded to initial instability growth and transition stages, with the remaining 4300 steps represent developed turbulence. Of these 4300 steps, 4000 were randomly selected for encoder training, and 4000 pairs of consecutive steps were used as a training set for a predictor of each tested model. Correspondingly, 300 time steps were reserved as a developer set to ensure no overfitting took place during training. This is a rather small training set, which represents a good test for model data-efficiency.

After training, each tested model was used to predict evolution of the system for 3000 additional time steps, using a single state of the system as an initial condition. Since HW turbulence is a chaotic system, reproducing it in exact agreement with the simulation is infeasible, and a comparison should be made for statistical characteristics. These include: 1) time-averaged spatial Fourier spectrum for both fields $n$ and $\phi$, 2) spatially-averaged temporal Fourier spectrums for both fields, 3) spatially and temporally averaged temporal auto-correlation for both fields, and 4) time-averaged numbers of local minima and maxima in the field of $\phi$ and the amplitude of their changes measured by the standard deviation. Since the system is periodic in both dimensions, time-averaged spatial Fourier spectra for both fields are circularly symmetric and can be presented as a 1D functions of radius (azimuthally integrated).

\subsection{The spectrum of tested models}

\begin{figure}[hbt!]
  \centering
  \includegraphics[width=1.0\textwidth]{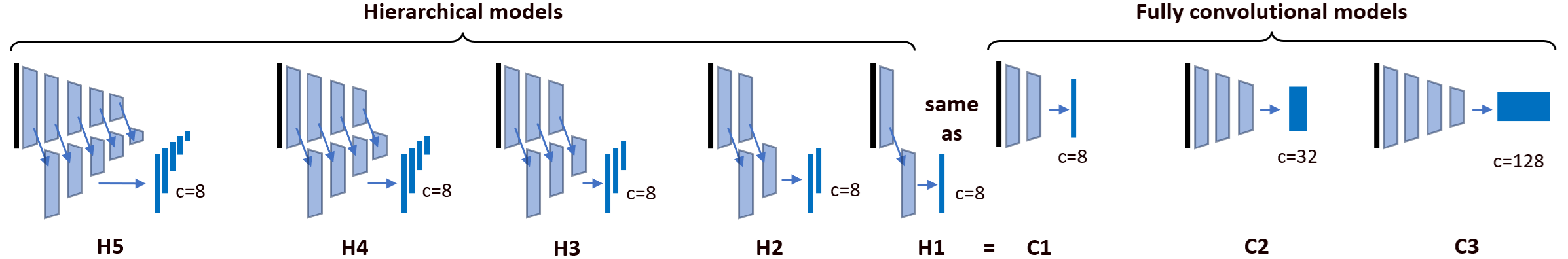}
  \caption{Variants of the models tested; c denotes of the number of embedding channels. Each trapezoid denotes a single convolutional layer with a 4x4 filter and stride 2.}
  \label{models}
\end{figure}

We compared the performance of several variants of our hierarchical model with various numbers of embedding layers against FCAE+ResNet models with encoders of various depth (see Fig. \ref{models}, where encoders of these models are shown). The input has a shape of 128x128x2, where the two channels are the $n$ and $\phi$ fields. 4x4 convolutions with stride 2 (preceded by a periodic padding) are used in each encoder’s layer, thereby, physical dimensions of each layer are reduced by a factor of two compared to the previous layer. In the deepest hierarchical model with 5 embedding layers (H5), physical dimensions of the deepest layer are just 2x2. Hierarchical models H1-H4 lack one or more deeper layers whereas shallow layers are the same as in the deepest model H5. The number of channels in each embedding layer is 8. A model with a single embedding layer (H1) is identical to a shallow fully convolutional model (C1), with the embedding of dimension 32x32x8. Other two convolutional models (C2 and C3) have correspondingly one and two more layers in the encoder (end symmetrically in the decoder) and an increased number of channels by factors of 4 and 16 respectively to keep the total number of embedding units the same. We have not included any DilResNet \cite{S2} models in the baseline for comparison because we experienced an order of magnitude slowdown even with just two (2-4-2) dilution residual blocks (similar number of parameres to C1 model).

\subsection{Comparing autoencoders}

\begin{wrapfigure}[21]{r}{0.51\textwidth}
  \vspace{-14pt}
  \centering
  \includegraphics[width=0.49\textwidth]{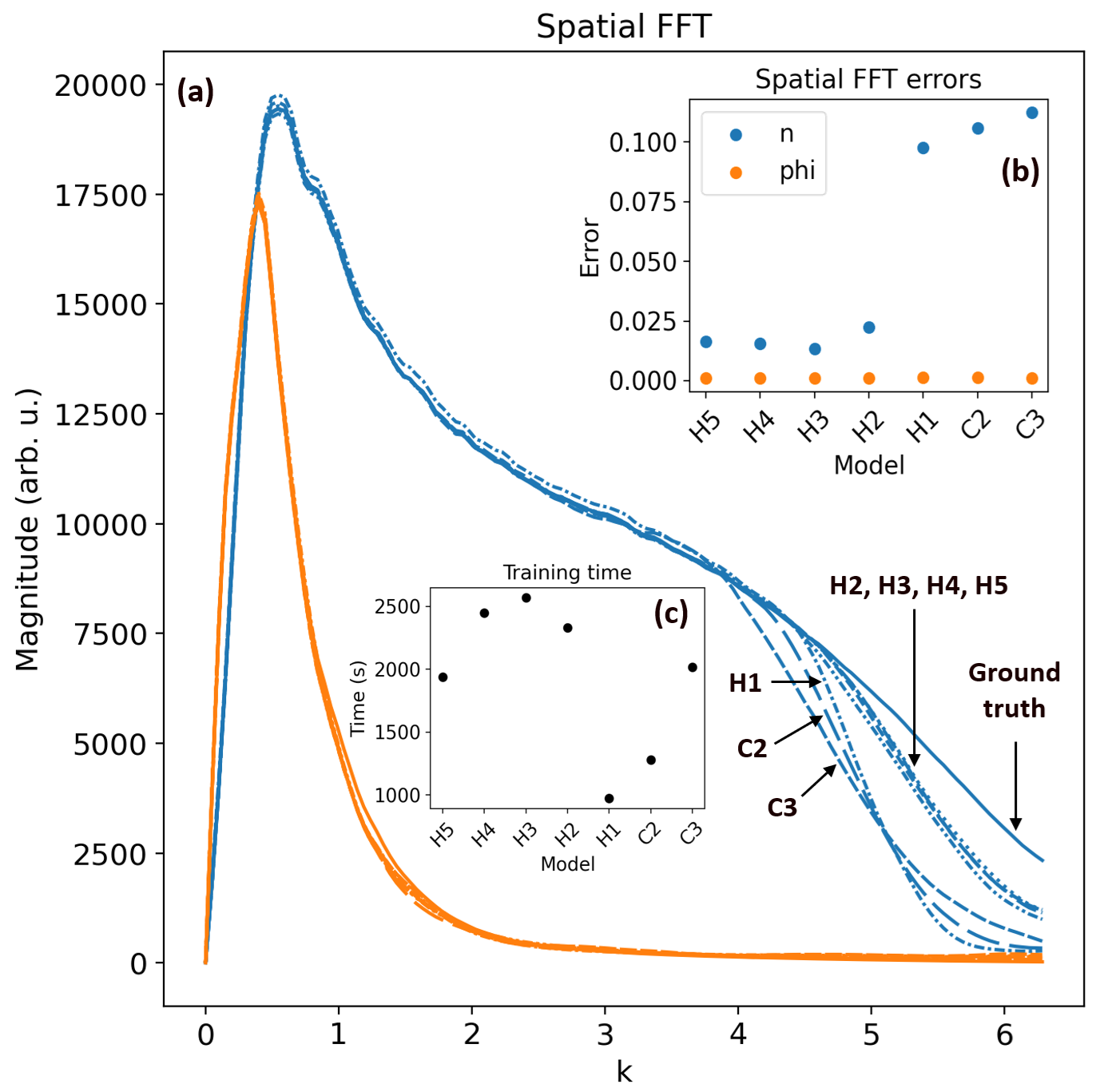}
  \caption{Comparing spatial spectra of recovered $n$ and $\phi$ fields with various autoencoders (a). Normalized overall deviation from the ground truth (b), and training time (c).}
  \label{AEcompare}
\end{wrapfigure}

We start by comparing the performance of autoencoders (without predictors), specifically their ability to accurately restore the input data. We analyze the accuracy of the models over the validation set, the results are presented in Fig. \ref{AEcompare} in the form of spatial FFT spectra for both input fields ($n$ and $\phi$) averaged over the 300 validation time steps (the same set of time steps was used in all models). The encoded 2D fields produced by all the autoencoders look virtually undistinguishable from the originals (by naked eye, see Fig. \ref{AEperf} in Appendix \ref{A_AEperf}).

Autoencoders of all models reproduce the field of $\phi$ accurately. The field of n is substantially more complex (see Fig. \ref{AEperf} in Appendix \ref{A_AEperf} and Figs. \ref{long_n}, \ref{long_phi} in Appendix \ref{long}), with the spectrum spanning a significantly higher frequency range (high $k$). All models reproduce most of the spectrum well, with some of high-frequency harmonics being under-resolved. For convenience, Fig. \ref{AEcompare}b shows overall deviations of the spatial FFT spectra from the ground truth (the spectrum of the input data) normalized by the variance of the ground truth data: $\text{error} = \overline{(y - y_{\text{true}})^2}/\operatorname{var}(y_{\text{true}})$. The same error definition was used in the following subsections. Notably, hierarchical models with at least 2 layers (H2-H5) outperform fully-convolutional models (C1=H1, C2, C3). Performance of the latter marginally deteriorates with the model’s depth as deeper models tend to slightly stronger over-smoothing of high-frequency details. 

The errors in the FFT spectra of recovered fields by all models are rather small. There is no need to improve the encoders, since sequential data generation for 3000 time steps ($\Delta t = 1$)  is a bigger source of error, as will be shown in the following subsections. The cost of training the hierarchical models is about a factor of two higher compared to the convolutional models (Fig. \ref{AEcompare}c).

\subsection{Comparing hierarchical models with different numbers of layers}

3000 time steps ($\Delta t = 1$)  is a sufficiently long rollout to test long-term prediction capabilities and obtain time-averaged quantities for analysis, since any small systematic error in the predictor, will accumulate over time.

\begin{figure}[hbt!]
  \centering
  \includegraphics[width=1.0\textwidth]{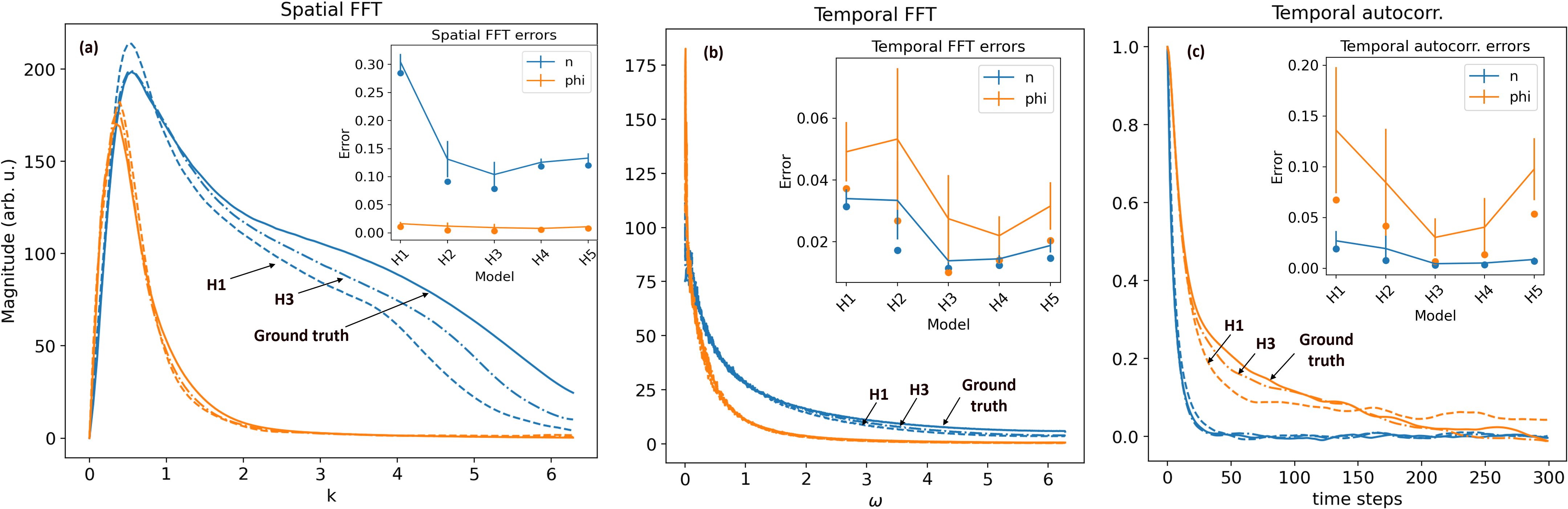}
  \caption{Comparing capabilities of the hierarchical models H1-H5 in long-sequence rollouts. Spatial (a) and temporal (b) FFT spectra, and auto-correlation (c) of the generated $n$ and $\phi$ fields. In the insets: dots show results by the best realization of each model, lines – mean and standard deviation for the top 6 realizations.}
  \label{Pcompare}
\end{figure}

\begin{figure}[hbt!]
  \centering
  \includegraphics[width=1.0\textwidth]{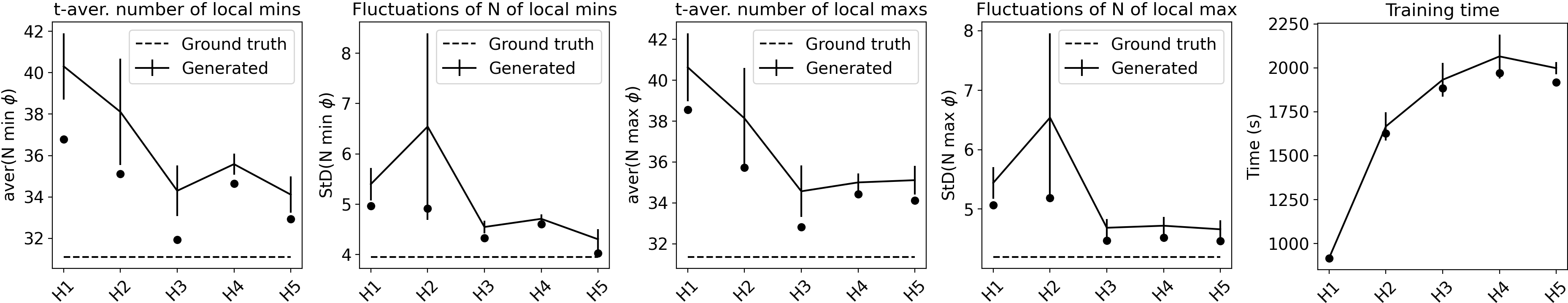}
  \caption{Comparing number of local minima and maxima in the field of $\phi$, which is a proxy for a number of vortices in the system, by H1-H5 models.}
  \label{Pcompare2}
\end{figure}

We noticed that realizations of the same predictor, trained using an identical procedure, but with different initializations can exhibit different long-term behavior, even if the training and validation losses have reduced at a similar rate during training. For a minority (10\%-20\%) of realizations the loss did not decrease. This behavior was also observed using the baseline models (C1-C3). It is not expected that all random initializations of a complex neural network yield the same training and rollout performance. For a systematic comparison of the models, we train 16 realizations of each predictor using different random initializations (the same set of embeddings is used, produced by a single realization of a corresponding AE). Fixed seeds \{2, 12, 22, 32, 42…\} are used to produce ‘random’ model initializations for reproducibility. We compare the models using three primary metrics: errors in spatial and temporal FFT spectra and temporal autocorrelations. We rank realizations by worst performing metric for each realization and select the top 6 realizations. For each metric of interest, we present the best result (shown by circles), the mean and the standard deviation of these 6 realizations, see Fig. \ref{Pcompare} (insets) and Fig. \ref{Pcompare2}. Predictors with N=4 internal steps (Fig. \ref{P}) were used in these tests.

The results can be interpreted as follows. If one trains a model using random initialization and wishes to achieve performance close to the best performing model shown here, roughly 10 to 20 attempts would be needed. To have a 90\% chance of obtaining a model with a performance within the range corresponding to the 6 best models plotted here, 5 attempts should suffice.

There is a clear improvement in performance (by all metrics) with the increase in number of embedding layers from 1 to 3, for both realization-averaged and best results. For the latter, errors reduce by a factor of 3 or more. The larger plots in Fig.\ref{Pcompare} show autocorrelations and spectra for higher performing realizations of the best (H3) and the worst model (H1). The spectra of $\phi$ are reproduced well by all models. The field of $n$ is significantly more complex, however the H3 model reproduces its spatial spectrum rather well, slightly underrepresenting high-frequency details. 

Importantly, an accurate spectrum of $\phi$ does not imply that all aspects of the field are reproduced correctly. The total numbers of local minima and maxima in $\phi$ are overpredicted by shallower models (Fig. \ref{Pcompare2}). These quantities can serve as a proxy for the number of vortices (emerging phenomena) in the system. A shallow model cannot capture long-range interaction between the vortices and allows them to come close to each other without due interaction.

The general shape of the temporal spectrum of both fields (Fig. \ref{Pcompare}b) is reproduced well by all models, although H3 is better at resolving higher frequencies. The biggest difference between the models is in the average level of $n$ and $\phi$ (0th harmonic) which is not accurate with shallower models resulting in higher overall error (see the inset). Notably, the field of $\phi$ is characterized by long-term auto-correlations (\ref{Pcompare}c) the capturing of which is a difficult task. Surprisingly, even a single-layer model H1 can capture this behavior reasonably, while the layered model H3 performs almost perfectly showing a strong improvement in the accuracy metric.

With deeper models (H4, H5), the performance does not improve further showing that the dominant scale of structures in the system is captured by the third level of representation. The number of embedding layers should be considered as a hyper-parameter of the model. 

As was the case for the autoencoders, the cost of training of the hierarchical predictors H2-H5 is about a factor of two higher compared to the convolutional model H1 (Fig. \ref{Pcompare2}), which is a small price to pay for a considerable improvement in accuracy.

\subsection{Baseline and ablation studies (see details in Appendices \ref{A_base} and \ref{A_abl})}

In Appendix \ref{A_base}, we demonstrate that the fully-convolutional model C1 (same as H1) represents a good baseline. Its performance cannot be improved either by increasing the number of internal steps N (Fig. \ref{P}) in the predictor from 4 in the standard model to 8 or 14, nor by increasing the encoder depth (models C2 and C3).

In Appendix \ref{A_abl}, we show that interactions between embedding layers in the predictor are crucial.

\section{Conclusion}

We have introduced the Hierarchical-Embedding Autoencoder-Predictor architecture for learning long-term time evolution in complex multi-scale 2D physical systems. It efficiently incorporates the scale hierarchy of features emerging within a physical system in its design. The hierarchical convolutional autoencoder transforms the state of a physical system into a series of embedding layers which encode structures of various scales preserving spatial information at a corresponding resolution level. The predictor advances all embedding layers in sync. In a case study of Hasegawa-Wakatani plasma turbulence, we have demonstrated that this architecture is capable of accurately predicting long-term evolution of this multi-scale system. We have analyzed physically-important statistical properties of the generated solutions and demonstrated that our model significantly outperforms conventional fully-convolutional (FCAE encoder + ResNet predictor) models which use a single embedding layer. Applying this architecture as a sub-grid model for fluid turbulence or a reduced-order model for kinetic turbulence could offer computational performance gains for device modeling.

\section{Acknowledgments}

This material is based upon work supported by the U.S. Department of Energy, Office of Science, under Award Number DE-SC0024522.

The authors are thankful to Álvaro Sánchez Villar (Princeton Plasma Physics Laboratory) for fruitful discussions.


\bibliographystyle{plainnat}  
\bibliography{lit}           

\newpage
\appendix

\section{Hasegawa-Wakatani turbulence model} \label{A_HW}
In this paper we have modeled plasma turbulence which is defined by a set of the Hasegawa-Wakatani (HW)  equations \cite{H83} relevant to fully-magnetized plasma in nuclear fusion devices. The model assumes that there is a gradient in the plasma density transverse to an external uniform magnetic field ${\bf B}=B {\bf b}_z$ (the field confinig the plasma in a tokamak or a stellarator). The ions are treated as cold and electrons have uniform temperature $T_e$. The HW equations consist of the continuity equations for electrons Eq.~(\ref{eq:ne}) and for ion guiding center density Eq.~(\ref{eq:ni}) together with the quasi-neutrality equation (\ref{eq:phi}) for electrostatic potential,
\begin{eqnarray}
&&\frac {\partial }{\partial t} \delta n_e+{\bf v}_{ExB}\cdot \nabla (n_0+\delta n_e)+n_0 {\bf b}_z\cdot\nabla v_{e\parallel}=0, \label{eq:ne} \\
&&\frac {\partial }{\partial t} \delta n_i^G+{\bf v}_{ExB}\cdot \nabla (n_0++\delta  n_i^G)=0, \label{eq:ni}  \\
&&n_e=n_i^G+n_0\frac {c}{B\omega_{ci}}\Delta \phi, \label{eq:phi}
\end{eqnarray}
where the last term in Eq.~(\ref{eq:phi}) is the ion polarization density $\delta n_i^{pol}=-n_0\nabla\cdot {\bf \xi}^{pol}$ with the ion polarization displacement ${\bf \xi}^{pol}=-(c/B\omega_{ci})\nabla \phi$. Here $n_0$ is the background plasma density, ${\bf v}_{E \times B}=(c/B) {\bf b}_z \times \nabla \phi$ is the particle drift velocity in the crossed electric ${\bf E}=-\nabla \phi$ and magnetic ${\bf B}$ fields, $\omega_{ci}=eB/m_ic$ is the ion cyclotron frequency, c is the speed of light, $m_i$ is the ion mass, and $e$ is the ion electric charge (hydrogen ions are assumed).
The electron velocity along the magnetic field $v_{e\parallel}$  in Eq.~(\ref{eq:ne}) is obtained from Ohm's law $\nu_{ei}v_{e\parallel}=(e/m_e){\bf b}_z\cdot\nabla \phi-(T_e/m_e){\bf b}_z\cdot\nabla \delta n_e/n_0$. Here $\nu_{ei}$ is the electron-ion collision frequency and $m_e$ is electron mass. The ion parallel velocity is negligible.  If we assume a single wavelength excited in the direction of the applied magnetic field with  the wavenumber $k_\parallel$ and introduce normalized quantities $e\phi/T_e=\bar \phi$, $\delta n_{e}/n_0=\bar n$ and normalized time $t\omega_{ci}=\bar t$ and space coordinates ${\bf x}/\rho_s=\bar  {\bf x}$, where $\rho_s^2=T_e/(m_i \omega_{ci}^2)$, the Eqs.~(\ref{eq:ne}--\ref{eq:phi}) can be rewritten in the dimensionless  form
\begin{eqnarray}
&&\frac {\partial }{\partial \bar t} \bar n +\{\bar\phi,\bar n\}+\kappa \frac {\partial}{\partial \bar y}\bar \phi=\alpha (\bar\phi-\bar n)-D_n \nabla^4 \bar n, \label{eq:n} \\
&&\frac {\partial }{\partial \bar t}\Delta \bar \phi+\{\bar \phi,\Delta \bar \phi\}=\alpha (\bar \phi-\bar n)-D_p \nabla^4 \bar \phi. \label{eq:p}
\end{eqnarray}
Here we assumed that the plasma gradient is in $x$ direction and  we defined two dimensionless  parameters $\kappa=-\rho_s d/d  x\log n_0$ and $\alpha=k_\parallel^2 T_e/(m_e\nu_{ei}\omega_{ci})>0$. The Poisson bracket in Eqs.~(\ref{eq:n}--\ref{eq:p}) is defined as $\{A, B\}=\frac {\partial A} {\partial x} \frac {\partial B}{\partial y}-\frac {\partial A} {\partial y} \frac {\partial B}{\partial x}$. We have also added two hyper-diffusive dissipative terms on the right hand side to model dissipative effects at short scales. These terms are necessary to achieve turbulent steady state in the simulations. 

When $\kappa>0$, Eqs.~(\ref{eq:n}--\ref{eq:p}) are subject to  the linear instability. As instability saturates, the nonlinear terms transfer energy from larger into smaller scales where it eventually dissipates by hyper-diffusivity. An additional dissipation is provided by parallel diffusivity proportional to $\alpha$.  The same nonlinear terms also transfer energy toward large scales where coherent strictures form (zonal flows) which help to regulate the turbulence.  For a given $\kappa$, the nature of the turbulence in the resulting statistical steady state depends on the parameter $\alpha$.  For $\alpha \ll 1$ equation~(\ref{eq:p})  decouples and becomes a 2D Euler equation. The density $\bar n$ is then almost passively advected by the ${\bf E}\times {\bf B}$ velocity \cite{Krommes2002}. 
In the opposite limit $\alpha \gg 1$, the  electron response is almost adiabatic $\bar \phi \approx \bar n$. Subtracting Eq.~(\ref{eq:p}) from Eq.~(\ref{eq:n})  leads to Hasegawa-Mima equations for $\bar \phi$ \cite{HM}. 

We have solved equations (\ref{eq:n}--\ref{eq:p}) for ${\bar n}$ and ${\bar \phi}$ (denoted $n$ and $\phi$ for simplicity in the rest of the paper) using the  BOUT++ \cite{B} code, for an intermediate range of parameters: $\alpha= 0.01$ and $\kappa = 0.5$. The hyper-diffusivity parameters were set to small values, $D_n = D_p = 0.0001$,  to ensure numerical stability without affecting the results. Computations were performed on a high-performance computing cluster utilizing eight A100 GPUs. The solver completed the task in approximately 3 hours. The solution (fields of $n$ and $\phi$ over 4800 time steps) are uploaded in the Supplementary Materials. 

\section{Model implementation details} \label{A_Model}

Details of the model will be provided upon completion of the paper review. In the meantime, inquiries can be directed to us via email akhrabry@princeton.edu

\subsection{Details of the hierarchical autoencoder model} \label{HAE_details}

\begin{figure}[hbt!]
  \centering
  \caption{Details of the model will be provided upon completion of the paper review. In the meantime, inquiries can be directed to us via email.}
  \label{OurAE}
\end{figure}

\subsection{Details of the autoencoder training procedure} \label{HAE_train}

\subsection{Details of the hierarchical predictor model} \label{P_details}

\begin{figure}[hbt!]
  \centering
  \caption{Details of the model will be provided upon completion of the paper review. In the meantime, inquiries can be directed to us via email.}
  \label{OurP}
\end{figure}

\subsection{Details of the predictor training procedure}\label{P_train}

\subsection{Computational complexity of the model} \label{m_complex}

\subsection{Accounting for spatial periodicity of the data}

\section{Baseline study: comparing fully-convolutional models of various depth} \label{A_base}

\begin{figure}[hbt!]
  \centering
  \includegraphics[width=1.0\textwidth]{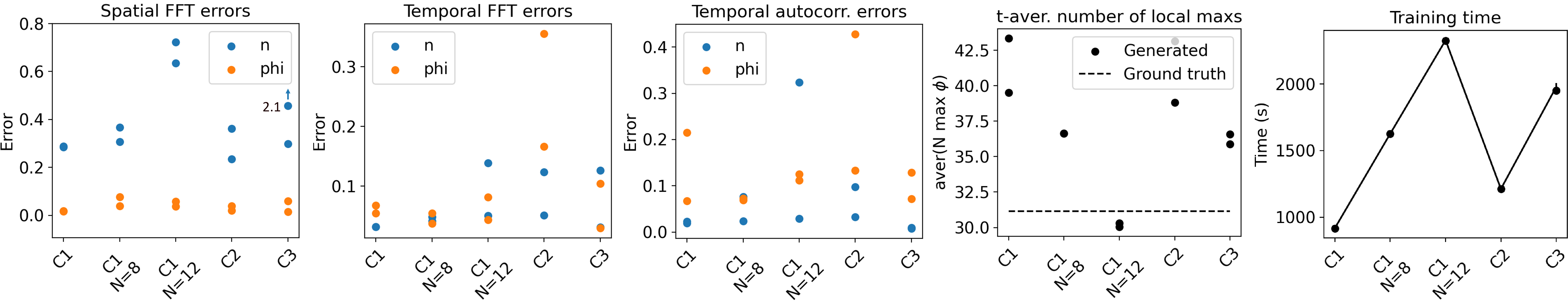}
  \caption{Comparing long sequence prediction errors and predictor training time of fully convolutional models with increasing depth of predictor (N=8, N=12) and autoencoder (C2, C3).}
  \label{basel}
\end{figure}

Here we demonstrate that the fully-convolutional model C1 (same as H1) represents a good baseline. Its performance cannot be improved neither by increasing the number of internal steps N (Fig. \ref{P}) in the predictor from 4 in the standard model to 8 or 14, nor by increasing the encoder depth (models C2 and C3). 10 realizations were trained of the C1 model with N=8 and N=14; and 12 realizations for each of the models C2 and C3 for which the learning rate was reduced by half for better stability of the training. The deeper models showed more irregular performance between realizations with some of the best 6 models showing very high errors, thereby we only present the results from the two best realizations for each model (Fig. \Ref{basel}). All deeper models resulted in higher errors than the shallow model C1 (which, in turn was worse than the hierarchical models H2-H5). The only exception is that the number of local maximums in the $\phi$ field was more accurately captured by the model C1 with a deep predictor N=12. However, this is likely coincidental since other aspects of the field are poorly captured by the model resulting in larger spatial FFT errors. Predictor training time for the deeper models was also larger by a factor of 1.5-2 (similar to the H2-H5 models).

\newpage
\section{Ablation studies} \label{A_abl}

\subsection{Removing interaction between embedding layers}

Here we show that interactions between embedding layers in the predictor are crucial.

\begin{figure}[hbt!]
  \centering
  \caption{Details of the model will be made available once the paper review is complete. For further information, please feel free to contact us via email}
  \label{P2}
\end{figure}

\newpage
\section{Long roll-out of H3 model (best realization)} \label{long}

\begin{figure}[hbt!]
  \centering
  \includegraphics[width=0.95\textwidth]{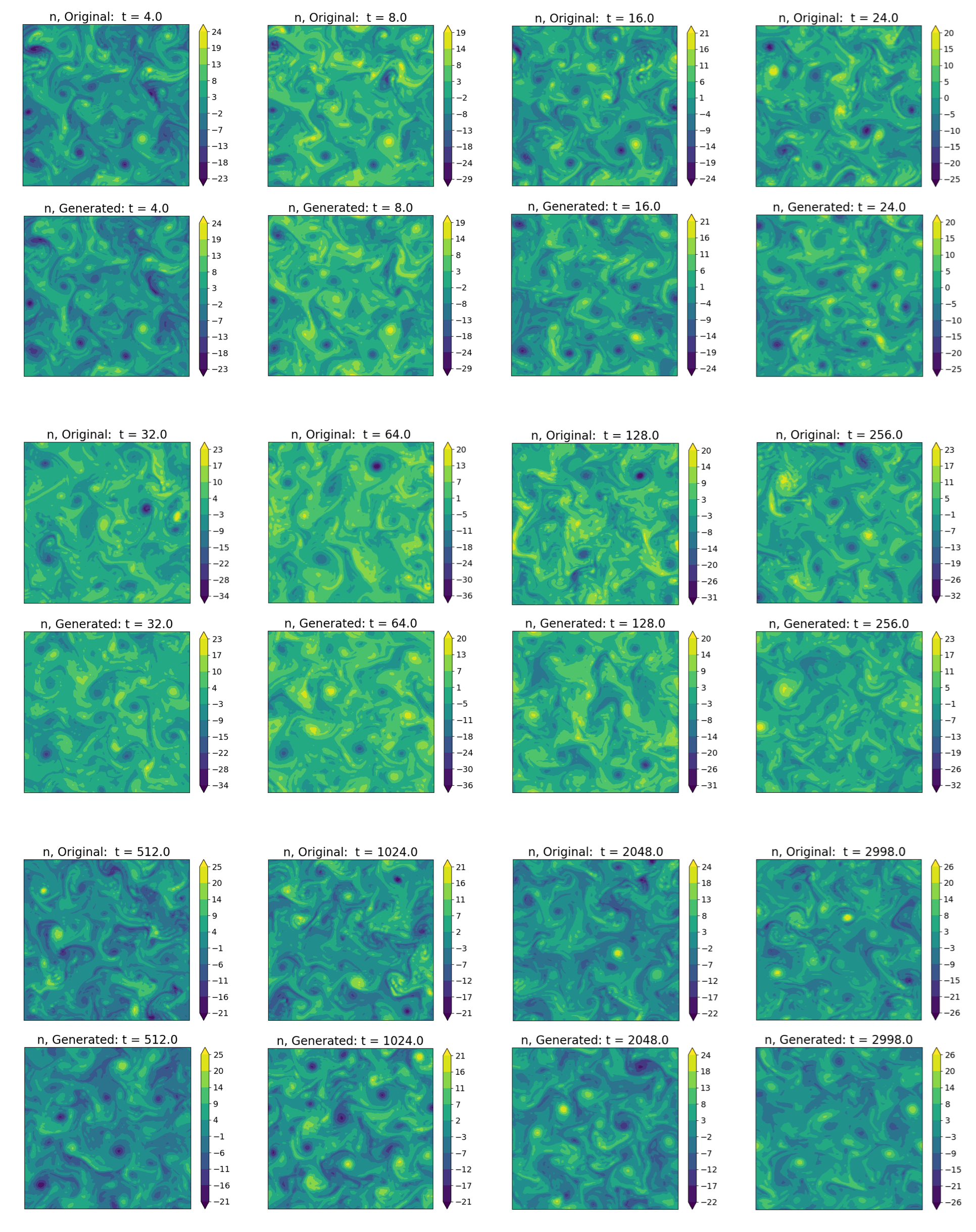}
  \caption{Long-term predictions of HW turbulence generated by the H3 model (best realization). The field of $n$: Generated data (using H3) VS numerical simulation (using BOUT++). At first, the generated solution resembles the original (simulated) one quite closely. However, with time, the differences amplify and by the time step 32 become significant. At later time steps, the solutions differ, however the pattern of the field is the same (as corroborated by statistical characteristics discussed in the paper). The full video is attached in supplementary materials.}
  \label{long_n}
\end{figure}

\hspace{0.1cm}
\newpage

\begin{figure}[hbt!]
  \centering
  \includegraphics[width=0.95\textwidth]{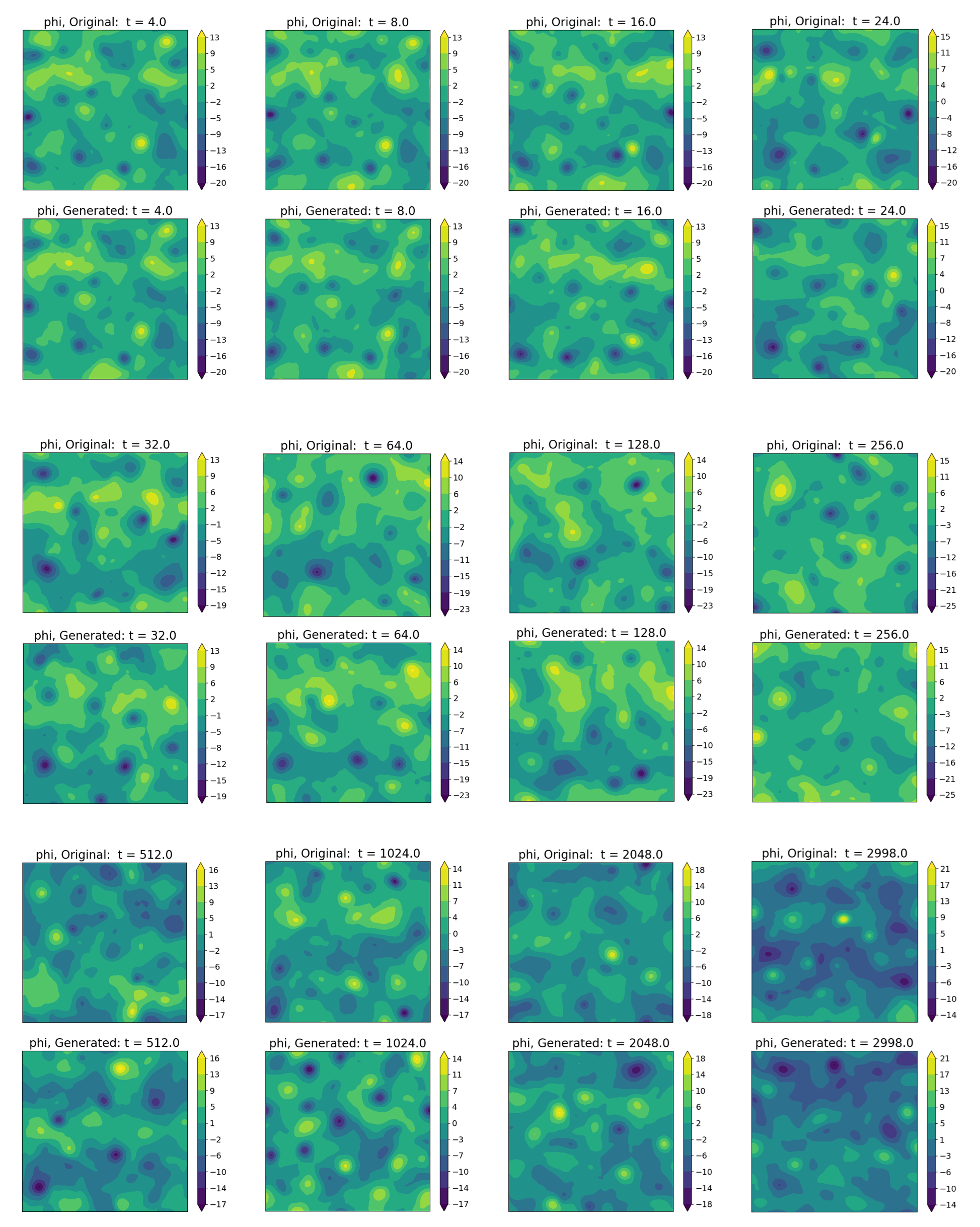}
  \caption{Long-term predictions of HW turbulence generated by the H3 model (best realization). The field of $\phi$: Generated data (using H3) VS numerical simulation (using BOUT++). Similarly to the field of $n$, at first, the generated solution resembles the original (simulated) one quite closely. However, with time, the differences amplify and by the time step 32 become significant. At later time steps, the solutions differ, however the pattern of the field is the same (as corroborated by statistical characteristics discussed in the paper). The full video is attached in supplementary materials.}
  \label{long_phi}
\end{figure}

\hspace{0.1cm}
\newpage

\section{Performance of the autoencoder of the H3 model} \label{A_AEperf}

\begin{figure}[hbt!]
  \centering
  \includegraphics[width=0.95\textwidth]{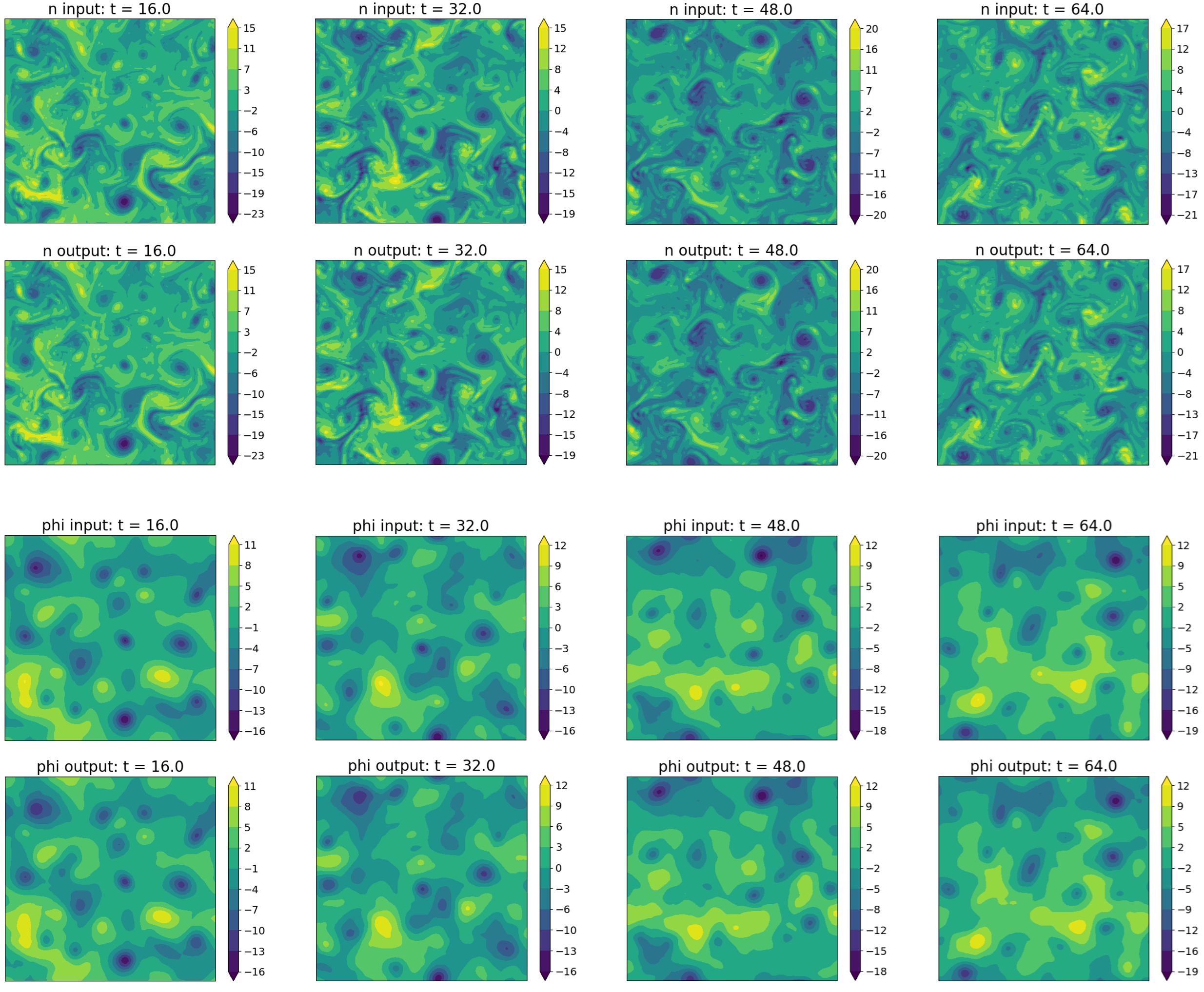}
  \caption{Comparing input and otput (recovered) fileds of $n$ and $\phi$ produced by AE of H3 for several time steps from the validation set. The outputs are indistinguishable (by naked eye) from the inputs.}
  \label{AEperf}
\end{figure}

\end{document}